\pdfoutput=1

\documentclass[11pt]{article}

\usepackage{acl}

\usepackage{times}
\usepackage{latexsym}

\usepackage[T1]{fontenc}

\usepackage[utf8]{inputenc}

\usepackage{microtype}

\usepackage{graphicx}               
\usepackage{tabularx}               
\usepackage{booktabs}
\usepackage{amsmath}
\usepackage{stmaryrd}
\usepackage{xcolor}
\usepackage{soul}

\newcolumntype{C}{>{\centering\arraybackslash}X}
\usepackage{multirow}               
\usepackage{diagbox}                
\usepackage{hhline}                 
\usepackage{color}                  
\usepackage{amsmath}                
\usepackage{amssymb}                
\usepackage{mathtools}              
\usepackage{enumitem}               

\usepackage{subfigure}
\usepackage{booktabs}
\usepackage{extarrows}
\usepackage{makecell}

\definecolor{RoseQuartzBg}{HTML}{F7CAC9}
\definecolor{RoseQuartz}{HTML}{F5A798}
\definecolor{Serenity}{HTML}{92A8D1}
\definecolor{OrangeRed}{rgb}{1.0, 0.27, 0.0}
\definecolor{Red}{rgb}{1.0, 0.0, 0.0}
\definecolor{Turquoise}{HTML}{0F4C81}
\usepackage{xparse}
\NewDocumentCommand{\lifu}{ mO{} }{\textcolor{OrangeRed}{\textsuperscript{\textit{Lifu}}\textsf{\textbf{\small[#1]}}}}
\NewDocumentCommand{\mo}{ mO{} }{\textcolor{blue}{\textsuperscript{\textit{Mo}}\textsf{\textbf{\small[#1]}}}}
\NewDocumentCommand{\minqian}{ mO{} }{\textcolor{blue}{\textsuperscript{\textit{Minqian}}\textsf{\textbf{\small[#1]}}}}
\NewDocumentCommand{\zhiyang}{ mO{} }{\textcolor{Red}{\textsuperscript{\textit{Zhiyang}}\textsf{\textbf{\small[#1]}}}}
\NewDocumentCommand{\sijia}{ mO{} }{\textcolor{Red}{\textsuperscript{\textit{Sijiia}}\textsf{\textbf{\small[#1]}}}}

\usepackage{graphics}

\title{
Incremental Prompting: Episodic Memory Prompt for Lifelong Event Detection

}

\author{Minqian Liu$^{\clubsuit}$, \ Shiyu Chang$^{\spadesuit}$, \ Lifu Huang$^{\clubsuit}$
\\
  $^{\clubsuit}$Virginia Tech, \ \ 
  $^{\spadesuit}$University of California Santa Barbara
 \\
  $^{\clubsuit}${\tt \{minqianliu,lifuh\}@vt.edu}, \ \ 
  $^{\spadesuit}${\tt chang87@ucsb.edu} 
  }

\begin{document}
\maketitle

\begin{abstract}

Lifelong event detection aims to incrementally update a model with new event types and data while retaining the capability on previously learned old types. One critical challenge is that the model would catastrophically forget old types when continually trained on new data. In this paper, we introduce \textbf{E}pisodic \textbf{M}emory \textbf{P}rompts (\textbf{EMP}) to explicitly retain the learned task-specific knowledge. Our method adopts continuous prompt for each task and they are optimized to instruct the model prediction and learn event-specific representation. The EMPs learned in previous tasks are carried along with the model in subsequent tasks, and can serve as a memory module that keeps the old knowledge and transferring to new tasks. Experiment results demonstrate the effectiveness of our method. Furthermore, we also conduct a comprehensive analysis of the new and old event types in lifelong learning.\footnote{The source code is publicly available at \url{https://github.com/VT-NLP/Incremental_Prompting}.}

\end{abstract}
\section{Introduction}

Class-incremental event detection~\cite{cao2020increment,yu2021lifelong} is a challenging setting in lifelong learning, where the model is incrementally updated on a continual stream of data for new event types while retaining the event detection capability for all the previously learned types. The main challenge of class-incremental event detection lies in the \emph{catastrophic forgetting} problem, where the model's performance on previously learned types significantly drops after it is trained on new data. Recent studies~\cite{lopez2017gradient,wang2019sentence} have revealed that replaying stored samples of old classes can effectively alleviate the catastrophic forgetting issue. However, simply fine-tuning the entire model on the limited stored samples may result in overfitting, especially when the model has a huge set of parameters. How to effectively leverage the limited stored examples still remains an important question.

Prompt learning, which is to simply tune a template-based or continuous prompt appended to the input text while keeping all the other parameters freezed, has recently shown comparable or even better performance than fine-tuning the entire model in many NLP tasks~\cite{gpt3,prefixtuning,ptuning}. It is especially flavored by lifelong learning since it only tunes a small amount of parameters. However, it is still non-trivial to equip prompts with the capability of retaining acquired knowledge and transferring to new tasks in the class-incremental setting.

In this work, we propose an incremental prompting framework that introduces \textbf{E}psodic \textbf{M}emory \textbf{P}rompts (\textbf{EMP}) to store and transfer the learned type-specific knowledge. At each training stage, we adopt a learnable prompt for each new event type added from the current task. The prompts are initialized with event type names and fine-tuned with the annotations from each task. To encourage the prompts to always carry and reflect type-specific information, we entangle the feature representation of each event mention with the type-specific prompts by optimizing its type distribution over them. After each training stage, we keep the learned prompts in the model and incorporate new prompts for next task. In this way, the acquired task-specific knowledge can be carried into subsequent tasks. Therefore, our EMP can be considered as a soft episodic memory that preserves the old knowledge and transfers it to new tasks. Our method does not require task identifiers at test time, which enables it to handle the challenging class-incremental setting. Our contributions can be summarized as follows:
\begin{itemize}[noitemsep,nolistsep,wide]
    \item We propose \textbf{E}psodic \textbf{M}emory \textbf{P}rompts (\textbf{EMP}) which can explicitly carry previously learned knowledge to subsequent tasks for class-incremental event detection. Extensive experiments validate the effectiveness of our method. 
    \item To the best of our knowledge, we are the first to adopt prompting methods for class-incremental event detection. Our framework has the potential to be applied to other incremental learning tasks.
\end{itemize}

\section{Problem Formulation}

Given an input text $x_{1:L}$ and a set of target spans $\{(x_i, x_j)\}$ from it, an event detection model needs to assign each target span with an event type in the ontology or label it as \textit{Other} if the span is not an event trigger. For class-incremental event detection, we aim to train a single model $f_{\theta}$ on a sequence of $T$ tasks $\{\mathcal{D}_1, ..., \mathcal{D}_T\}$ that consist of non-overlapping event type sets $\{\mathcal{C}_1, ..., \mathcal{C}_T\}$\footnote{Though the type sets from all tasks contain \textit{Other}, they have distinct meanings given different seen types.}. In each $t$-th task, the model needs to classify each mention to any of the types that have seen so far $\mathcal{O}_t=\mathcal{C}_1\bigcup...\bigcup\mathcal{C}_t$. The training instances in each task $\mathcal{D}_t$ consist of tuples of an input text $x_{1:L}^{t}$, a target span $\bar{x}^{t}$, and its corresponding label $\mathbf{y}^t$ where $\mathbf{y}^t\in\mathcal{C}_t$. For convenience, the notations are for the $t$-th training stage by default unless denoted explicitly in the following parts of the paper.

\section{Approach}

\subsection{Span-based Event Detection}
\label{sec:baseline}
Given an input sentence $x_{1:L}^t$ from task $\mathcal{D}_t$, we first encode it with BERT~\cite{bert} to obtain the contextual representations $\mathbf{x}_{1:L}^t = \text{BERT}(x_{1:L}^t)$. Note that we freeze BERT's parameters in our method and all baselines. For each span $\bar{x}^{t}$, we concatenate its starting and ending token representations and feed them into a multilayer perceptron (MLP) to get the span representation $\mathbf{h}_{span}^{t}$. Then, we apply a linear layer on $\mathbf{h}_{span}^{t}$ to predict the type distribution of the span $p^{t}=linear(\mathbf{h}_{span}^{t})$. We use cross-entropy loss to train the model on $\mathcal{D}_t$:
\begin{equation}
\mathcal{L}_{C} = -\sum_{(\bar{x}^{t},y^t)\in \mathcal{D}_t} \text{log}\ p^{t}.
\label{equa:span_detection}
\end{equation}

\subsection{Episodic Memory Prompting} 
\label{sec_emp}
To overcome the catastrophic forgetting and exemplar memory overfitting issues, we design an incremental prompting approach with Episodic Memory Prompts (EMPs) to preserve the knowledge learned from each task and transfer to new tasks.

Given an incoming task $\mathcal{D}_t$ and its corresponding new event type set $\mathcal{C}_t = \{c_1^t,...,c_{n_t}^t\}$, we first initialize a sequence of new \emph{prompts} $\mathbf{C}^t=[\mathbf{c}_1^t,...,\mathbf{c}_{n_t}^t]$ where $\mathbf{c}_i^t\in\mathbb{R}^{1\times e}$ is a type-specific prompt for type $c_i^t$, $n_t$ is the number of event types in the $t$-th task. $e$ is the embedding dimension size. In our experiments, we use the event type name to initialize each event type prompt $\mathbf{c}_i^t$ (see Appendix~\ref{sec:implement} for details). Note that we always preserve the prompts learned from previous tasks, thus the accumulated prompts until the $t$-th task are represented as $\mathbf{I}^t=[\mathbf{C}^1,...,\mathbf{C}^t]$. Given a particular sentence $x_{1:L}^t$ from $\mathcal{D}_t$, we concatenate it with the accumulated prompts $\mathbf{I}^t$, encode the whole sequence with BERT, and obtain the sequence of contextual representations $[\tilde{\mathbf{x}}_{1:L}^t; \tilde{\mathbf{I}}^t]$, where $\tilde{\mathbf{x}}_{1:L}^t$ and $\tilde{\mathbf{I}}^t$ denote the sequence of contextual embeddings of $x_{1:L}^t$ and $\mathbf{I}^t$ respectively. $[;]$ is concatenation operation. Then, similar as Section~\ref{sec:baseline}, we obtain a representation $\tilde{\mathbf{h}}_{span}^{t}$ for each span based on $\tilde{\mathbf{x}}^t_i$, and predict the logits over all target event types $\tilde{p}^{t}=linear(\tilde{\mathbf{h}}_{span}^{t})$. 

We expect the EMPs to be specific to the corresponding event types and preserve the knowledge of each event type from previous tasks. So we design an entangled prompt optimization strategy to entangle the feature representation of each span with the event type-specific prompts by computing an event type probability distribution over them. Specifically, given a span representation $\tilde{\mathbf{h}}_{span}^{t}$ and EMP representations $\tilde{\mathbf{I}}^t$, we compute the probability distribution over all prompts as $\tilde{p}^t_c = \text{MLP}(\tilde{\mathbf{I}}^t)\cdot\tilde{\mathbf{h}}_{span}^{t}$, where $\cdot$ is the dot product. Finally, we combine the original logits $\tilde{p}^{t}$ and $\tilde{p}^t_c$ to predict the event type label for each span:
\begin{equation}
\label{equa:emp}
\tilde{\mathcal{L}}_{C} = -\sum_{(\bar{x}^t,y^t)\in \mathcal{D}_t} \text{log}\ (\tilde{p}^{t} + \tilde{p}^t_c)
.
\end{equation}

At the end of each training stage, we keep the learned prompts from the current task $\mathbf{C}^t$ in the model, and then initialize a new prompt $\mathbf{C}^{t+1}$ for the next task and concatenate it with the previous accumulated prompts $\mathbf{I}^{t}$ incrementally: $\mathbf{I}^{t+1} = [\mathbf{I}^{t};\mathbf{C}^{t+1}]$.

\subsection{Lifelong Learning with Experience Replay and Knowledge Distillation}
To alleviate the catastrophic forgetting issue, two strategies have been widely applied in many lifelong learning works~\cite{icarl,sun-etal-2020-distill,cao2020increment,yu2021lifelong}: (1) Experience Replay which is to repeatedly optimize the model on the stored previous data in subsequent tasks; and (2) Knowledge Distillation (KD) that is to ensure the output probabilities and/or features from the current and previous models to be matched, respectively. We also adopt these two baselines to validate the compatibility of our method with other lifelong learning techniques.

Specifically, after training on $\mathcal{D}_t$, we apply the herding algorithm~\cite{herding} to select 20 training samples for each type into the memory buffer, denoted as $\mathcal{M}$. Similar as Equation~\ref{equa:emp}, the objective for experience replay is:
\begin{equation}
\mathcal{L}_{ER} = -\sum_{(\bar{x}^r,y^r)\in \mathcal{M}} \text{log}\ (\tilde{p}^{t} + \tilde{p}^t_c).
\end{equation}

For knowledge distillation, following~\cite{cao2020increment}, we apply both \emph{prediction-level} and \emph{feature-level} distillation.
The objectives for prediction-level KD and feature-level KD are computed as:
\begin{equation*}
    \mathcal{L}_{PD} = - \sum_{(\bar{x}^r, y^r)\in\mathcal{M}} (\tilde{p}^{t-1} + \tilde{p}^{t-1}_c)\ \text{log}\ ((\tilde{p}^{t} + \tilde{p}^t_c)),
\end{equation*}
\begin{equation*}
    \mathcal{L}_{FD} = \sum_{(x^r, (x_i^r, x_j^r),y^r)\in\mathcal{M}} 1 - g(\Bar{\mathbf{h}}_{span}^{t-1}, \Bar{\mathbf{h}}_{span}^{t}),
\end{equation*}
where $g$ is the cosine similarity function. $\Bar{\mathbf{h}}_{span}^{t-1}$ and $\Bar{\mathbf{h}}_{span}^{t}$ are $l_2$-normalized features from the model at $t-1$ and $t$ stages, respectively.

\paragraph{Optimization}
We combine the multiple objectives with weighting factors $\alpha$ and $\beta$ as follows:
\begin{equation*}
    \mathcal{L} = \tilde{\mathcal{L}}_C + \alpha\mathcal{L}_{ER}+\beta(\mathcal{L}_{PD}+\mathcal{L}_{FD}).
    \label{equa:baseline_train}
\end{equation*}

\section{Experiments and Discussions}

\paragraph{Experiment Settings} We conduct experiments on two benchmark datasets: ACE05-EN~\cite{ace} and MAVEN~\cite{maven}, and construct the class-incremental datasets following the \emph{oracle negative} setting in~\cite{yu2021lifelong}. We divided the ontology into 5 subsets with distinct event types, and then use them to constitute a sequence of 5 tasks denoted as $\mathcal{D}_{1:5}$. We use the same partition and task order permutations in \cite{yu2021lifelong}. During the learning process from $\mathcal{D}_1$ to $\mathcal{D}_5$, we constantly test the model on the entire test set (which contains the whole ontology) and take the mentions of unseen event types as negative instances. More implementation details, including parameters, initialization of prompts as well as baselines are shown in Appendix~\ref{appendix_exp}.

\paragraph{Baselines} We consider the following baselines for comparison: (1) \textbf{BERT-ED}: simply trains the BERT based event detection model on new tasks without prompts, experience replay or knowledge distillation. It's the same as the span-based event detection baseline in Section~\ref{sec:baseline}. (2) \textbf{KCN}~\cite{cao2020increment}: use a prototype-based example sampling strategy and hierarchical distillation. As the original approach studied a different setting, we adapt their prediction-level and feature-level distillation as the baseline. (3) \textbf{KT}~\cite{yu2021lifelong}: transfer knowledge between old types and new types in two directions. (4) \textbf{iCaRL*}~\cite{icarl}: use nearest-mean-of-exemplars rules to perform classification combined with knowledge distillation. iCaRL adopts different strategies for classification, experience replay, and distillation. We thus directly report the result in~\cite{yu2021lifelong} for reference. (5) \textbf{EEIL}~\cite{eeil}: use an additional finetuning stage on the balanced dataset. (6) \textbf{BIC}~\cite{bic}: use a bias correction layer after the classification layer. (7) \textbf{Upperbound}: trains the same model on all types in the datasets jointly. 
For \textbf{iCaRL}, \textbf{EEIL}, and \textbf{BIC}, we use the same implementation in~\cite{yu2021lifelong}.
For fair comparison, our approach and all baselines (except for the Upperbound baseline) are built upon \textbf{KCN} and use the same experience replay and knowledge distillation strategies described in Section~\ref{sec_emp}. We set the exemplar buffer size as 20, and allow one exemplar instance to be used in each training batch instead of the whole memory set. Note that this replay setting is different from the one in~\cite{yu2021lifelong}, where we allow much less frequent exemplar replay, and thus our setting is more efficient, challenging, and realistic.

\begin{table*}[!htbp]
	\centering
	\resizebox{0.9\textwidth}{!}
	{
	\begin{tabular}{l | ccccc | ccccc}
    \toprule
    & \multicolumn{5}{c}{MAVEN}  & \multicolumn{5}{c}{ACE05-EN} \\
    \midrule
    Task &1 &2 & 3 & 4 & 5 & 1 &2 & 3 & 4 & 5  \\ 
    \midrule
    BERT-ED & 63.51 & 39.99 & 33.36 & 23.83 & 22.69 & 58.30 & 43.96 & 38.02 & 21.53 & 25.71 \\
    iCaRL*~\cite{icarl} & 18.08 & 27.03 & 30.78 & 31.26 & 29.77 & 4.05 & 5.41 & 7.25 & 6.94 & 8.94 \\
    EEIL~\cite{eeil} & 63.51 & 50.62 & 45.16 & 41.39 & 38.34 & 58.30 & 54.93 & 52.72 & 45.18 & 41.95 \\
    BIC~\cite{bic} & 63.51 & 46.69 & 39.15 & 31.69 & 30.47 & 58.30 & 45.73 & 43.28 & 35.70 & 30.80 \\
    KCN~\cite{cao2020increment} & 63.51 & 51.17 & 46.80 & 38.72 & 38.58 & 58.30 & 54.71 & 52.88 & 44.93 & 41.10 \\
    KT~\cite{yu2021lifelong} & 63.51 & 52.36 & 47.24 & 39.51 & 39.34 & 58.30 & \textbf{55.41} & 53.95 & 45.00 & 42.62 \\
    \midrule
    EMP (Ours) &\textbf{67.86} & \textbf{60.26} & \textbf{58.61} & \textbf{54.81} & \textbf{50.12} & \textbf{59.60} & 53.19 & \textbf{55.20} & \textbf{45.64} & \textbf{43.28} \\ 
    \midrule
    Upperbound (Ours) & / & / & / & / & 68.42 & / & / & / & / & 67.22 \\
    \bottomrule
    \end{tabular}
	} 
	\caption{Comparison between our approach and baselines in terms of micro F-1 (\%) on 5 class-incremental tasks. We report the \emph{averaged} results on 5 permutations of tasks so that the results are independent of randomness.}
	\vspace{-1.2em}
	\label{tab:main}
\end{table*}

\begin{figure}[tbp]
	\centering
	\includegraphics[width=0.7\linewidth]{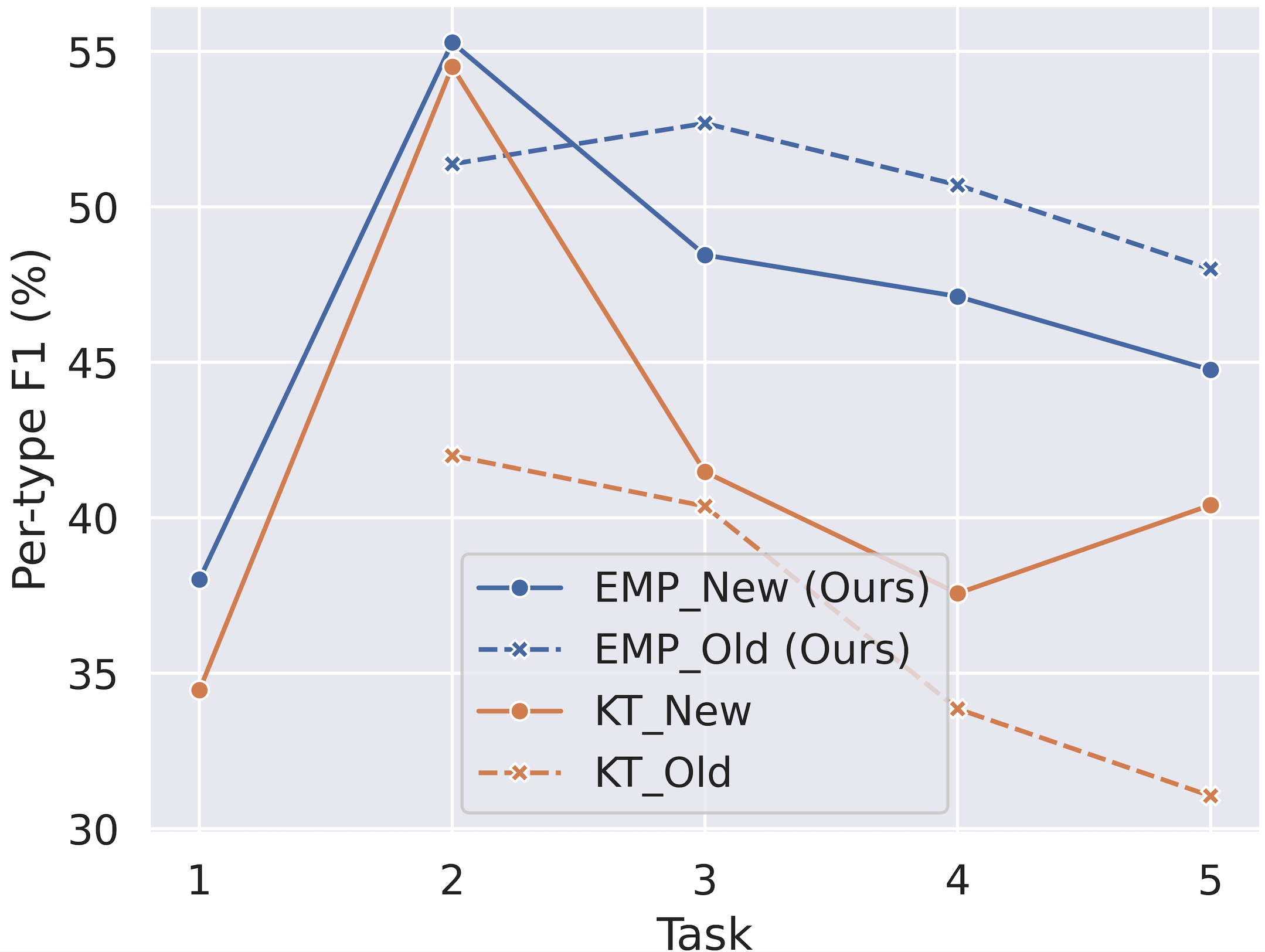}
	\caption{Per-type F1 on old types and new types in each lifelong task on one \emph{randomly selected} permutation of the MAVEN dataset. The F-scores on old and new types reflect the ability to retain acquired knowledge and to learn new types, respectively. Best viewed in color.}
	\vspace{-1.2em}
	\label{fig:overall}
\end{figure}

\paragraph{Results}
We present the main results in Table~\ref{tab:main}. We have following observations: 
(1) by comparing the performance of various approaches on Task 1 which are not affected by any catastrophic forgetting, our approach improves 4.1\% F-score on MAVEN and 1.3\% F-score on ACE05, demonstrating that by incorporating task-specific prompts, event detection itself can be significantly improved. EMPs even provide more improvement on MAVEN which contains a lot more event types than ACE05, suggesting the potential of incorporating EMPs for fine-grained event detection; (2) \textbf{KCN} can be viewed as an ablated version of our approach without EMPs. Our approach consistently outperforms \textbf{KCN} on almost all tasks on both datasets, demonstrating the effectiveness of EMPs on improving class-incremental event detection; (3) Comparing with \textbf{BERT-ED}, \textbf{KCN} adopts experience replay and knowledge distillation. Their performance gap verifies that these two strategies can dramatically alleviate catastrophic forgetting; (4) There is still a large gap between the current approaches and the upperbound, indicating that catastrophic forgetting still remains a very challenging problem. Note that the only difference in \textbf{EEIL}, \textbf{BIC}, \textbf{KCN}, and \textbf{KT} is the lifelong learning techniques they applied, thus these models have identical F-score on Task 1. We also analyze failed examples in Appendix~\ref{appendix_fail}.
\paragraph{Analysis of New and Old Types in Lifelong Learning} Figure~\ref{fig:overall} shows the F-score on old and new event types in each training stage for our approach and \textbf{KT}~\cite{yu2021lifelong} on MAVEN. Our approach consistently outperforms \textbf{KT} by a large margin on both old types and new types, demonstrating that our EMPs effectively preserve learned knowledge from old event types and improve event detection when annotations are sufficient. Interestingly, comparing the F-score on new types in Task 1 and old types in Task 2, both methods improve the performance on the types of Task 1, indicating that both methods have the potential of leveraging indirect supervision to improve event detection.

\paragraph{Ablation Study}
We consider four ablated models based on our EMPs: (1) change the prompt initialization\footnote{Appendix~\ref{appendix_exp} shows the details of prompt initialization. We use the same initialization for the discrete prompt ablation.} from using event type name representations
to using random distribution; (2) remove the entangled prompt optimization but still append the event type prompts to the end of each input sentence and apply Equation~\ref{equa:span_detection} only to detect the events; (3) remove the knowledge distillation loss $\mathcal{L}_{PD}$ and $\mathcal{L}_{FD}$; (4) use completely fixed prompts to replace the trainable soft prompts. From Table~\ref{tab:ablation}, we observes that: (1) using event type names to initialize the prompts is helpful in most tasks; (2) both entangled prompt optimization and knowledge distillation can help alleviate catastrophic forgetting; (3) switching the continuous prompts to discrete prompts degrades the performance significantly, suggesting that the continuous prompts are generally more promising than discrete prompts. 

\begin{table}[!htbp]
	\centering
	\resizebox{0.45\textwidth}{!}
	{
		\begin{tabular}{c  ccccc}
    \toprule
    Task &1 &2 & 3 & 4 & 5  \\ 
    \midrule
    EMP (Ours) & \textbf{67.86} & \textbf{60.26} & \textbf{58.61} & \textbf{54.81} & \textbf{50.12} \\
    - w/o EInit & 66.73 & 58.99 & 57.63 & 53.98 & 49.33 \\
    - w/o EPO & 67.04 & 59.02 & 57.79 & 53.72 & 49.05 \\
    - w/o KD & 67.86 & 57.57 & 55.83 & 53.02 & 48.65 \\
    - Discrete & 60.13 & 51.98 & 50.60 & 48.97 & 43.68 \\
    \bottomrule
    \end{tabular}
	} 
	\caption{Ablation study on event-specific prompt initialization (EInit), entangled prompt optimization (EPO), knowledge distillation (KD), and trainable soft prompts (Discrete) on MAVEN. We report the \emph{averaged} results on 5 permutations of tasks.}
	\vspace{-1em}
	\label{tab:ablation}
\end{table}

\begin{figure}[tbp]
	\centering
	\includegraphics[width=0.7\linewidth]{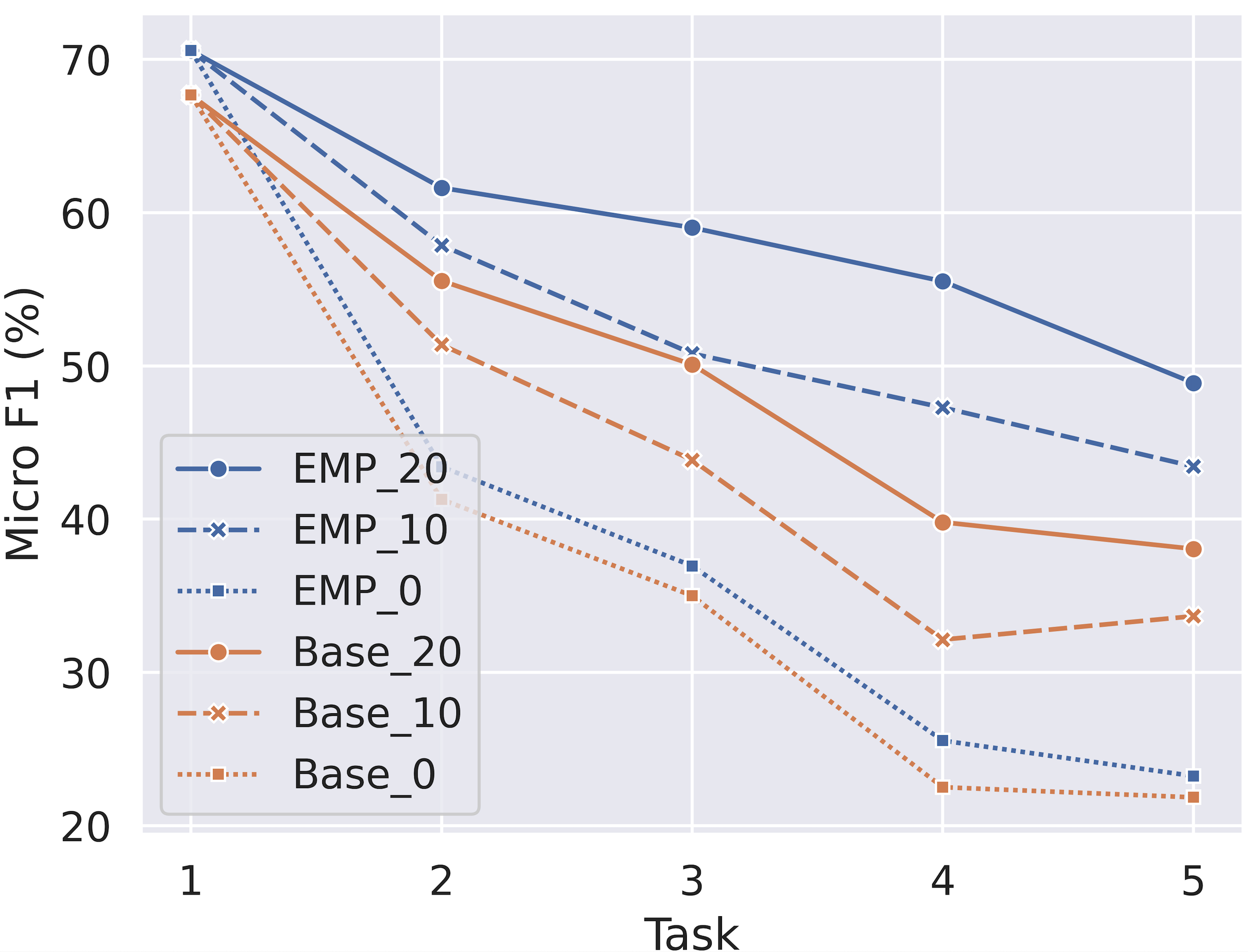}
	\caption{Performance with different buffer size in each task on one \emph{randomly selected} permutation of MAVEN. Best viewed in color.}
	\label{fig:buffer}
\end{figure}

\paragraph{Effect of Exemplar Buffer Size}
\label{appendix_effect}
We conduct an analysis on the effect of exemplar buffer size. We explore the buffer size for each type in \{0, 10, 20\}. 
We use \textbf{KT} as the baseline when buffer size is 20 and 10. Note that when buffer size is 0, we do not adopt either experience replay or knowledge distillation and thus use \textbf{BERT-ED} as the baseline. We plot the results on Figure~\ref{fig:buffer}. 
We observed that: (1) Decreasing the buffer size for each type from 20 to 10 degrades the performance of both models. This indicates that reducing data diversity may result in the overfitting on example data, and thus deteriorates the performance; 
(2) Our method still outperforms the \textbf{KT} baseline when storing only half of history examples, which indicates our method is able to utilize the stored examples more effectively.
(3) When the buffer size decreases to 0, the performance of both methods drops significantly. This shows that both approaches highly rely on the stored data to overcome the catastrophic forgetting problem. This calls for developing more advance techniques to reduce the dependence on stored examples, as storing past data could result in data leakage in real-world applications.

\section{Related Work}
\paragraph{Lifelong Event Detection} Deep neural networks have shown state-of-the-art performance on supervised event detection~\cite{jointEE,feng2016,zhang2017improving,huang2020semi,wang2021query}. However, when moving to lifelong learning setting, their performance significantly drops~\cite{kirkpatrick2017overcoming,aljundi2019gradient,biesialska-etal-2020-continual,li2021refining,ke-etal-2021-adapting,madotto-etal-2021-continual,ke-etal-2021-classic,feng2022hierarchical}. Though experience replay~\cite{lopez2017gradient,dautume2019episodic,guo2020improved,han2020continual,zhao2022consistent} and knowledge distillation~\cite{llkd,cao2020increment} have shown to be effective in overcoming catastrophic forgetting, they highly rely on the stored data from old tasks, which is not the most realistic setting for lifelong learning.

\paragraph{Prompt Learning}

Conditioning on large-scale pre-trained language models, prompt learning~\cite{gpt3,ptuning,ptuningv2,transprompt,l2p,wang2022art} has shown comparable performance as language model fine-tuning. Specific to lifelong learning, \citet{lept5} use prompt tuning to train the model as a task solver and data generator for lifelong few-shot problem. \citet{zhu2022CPT} propose continual prompt tuning for dialogue state tracking. To the best of our knowledge, we are the first work to adopt prompt learning for class-incremental event detection. 


\section{Conclusion}

We propose a novel Episodic Memory Prompting (EMP) framework for class-incremental event detection. During each training stage, EMP learns type-specific knowledge via a continuous prompt for each event type. The EMPs trained in previous tasks are kept in the model, such that the acquired task-specific knowledge can be transferred into the following new tasks. Experimental results validate the effectiveness of our method comparing with competitive baselines. Our extensive analysis shows that by employing EMPs, both event detection itself and the incremental learning capability of our approach are significantly improved.  


\section*{Acknowledgements}
We thank the anonymous reviewers and area chair for their valuable time and constructive comments, and the helpful discussions with Zhiyang Xu. We also thank the support from the Amazon Research Awards. 





\bibliography{anthology,custom}
\bibliographystyle{acl_natbib}

\appendix

\renewcommand{\arraystretch}{0.5}

\section{Experimental Details}
\label{appendix_exp}

\paragraph{Implementation Details}
\label{sec:implement}
During training, we use AdamW~\cite{adamw} optimizer with the learning rate set to $1e-4$ and weight decay set to $1e-2$. Different from previous work~\cite{yu2021lifelong}, we set the batch size to 1 as we encode each sentence once and consider all target spans in the sentence at the same time. We adopt gradient accumulation with the step set to 8. As the number of batches is large, we apply a periodic replay and distillation strategy with the interval set to 10 to reduce computational cost. For each lifelong task $\mathcal{D}_{t}$, we set the maximum number of training epochs to 20. We adopt the early stopping strategy with patience 5, i.e., the training stops if the performance on the development set does not increase for 5 epochs. 
The temperature parameter used in prediction-level distillation is set to 2. The weighted factors for the loss function $\alpha$ and $\beta$ are computed based on the number of learned event types and new types.

The parameters of each prompt in EMPs are initialized with the corresponding event type name. Specifically, there are three cases in the initialization: (1) If the type name is \emph{single-token} and it is contained in BERT's vocabulary, we directly use the pre-trained embedding of this token to initialize the prompt; (2) If the type name is \emph{multiple-token} and the tokens are contained in BERT's vocabulary, we take the average of the pre-trained embeddings of these tokens to initialize the prompt; (3) If the type name contains \emph{Out-of-Vocabulary (OOV)} tokens, we replace the OOV tokens with the synonyms that are contained in BERT's vocabulary. It is worth noting that we randomly initialize the prompt for the \emph{Other} type and keep updating it throughout all lifelong tasks. We leave how to incorporate more effective prior knowledge into prompts for future work.

\section{Failure Cases}
\label{appendix_fail}
We show some of typical failure cases in Table~\ref{tab:case_study}. We have following observations: (1) the first three examples illustrate the catastrophic forgetting problem in class-incremental event detection. While the model predicted correct event types right after it was trained on those types, it starts to predict wrong types in subsequent tasks. Interestingly, we observed that the model typically predicts the \emph{Other} type or the types relevant to triggers (e.g., \emph{Creating}) when forgetting occurs; (2) the 4th and 5th examples showed that the model sometimes keeps predicting the old types while it is supposed to predict new types in subsequent tasks. (3) the 7th example showed that the model can sometimes correct itself in subsequent tasks, which indicates the experience replay and knowledge distillation have the potentials of improving old types; (4) the last example indicates that in some cases, the model is interfered after trained on a task contained ambiguous types even though it predicts the correct type in all other tasks.

\begin{table*}[!htbp]
	\centering
	\resizebox{1\textwidth}{!}
	{
		\begin{tabular}{m{0.4\textwidth}>
		{\centering}m{0.2\textwidth}>
		{\centering\arraybackslash}m{0.4\textwidth}}
    \toprule
    \multicolumn{1}{c}{Text} & Gold Event Type(s) & Predicted Event Type(s)  \\ 
    \midrule
    The Minnesota Territory itself was \textbf{formed} only in 1849 but the area had a rich history well before this. & Coming\_to\_be ($\mathcal{D}_2$) & \underline{$f_2$}: Coming\_to\_be; $f_{3:4}$: Other; \hspace{15 mm} $f_5$: Creating ($\mathcal{D}_5$) \\
    \midrule
    He \textbf{informed} the Air France chief executive in writing "I did not believe the captain capable of qualifying in the 707." & Telling ($\mathcal{D}_3$) & \underline{$f_3$}: Telling; $f_4$: Other; $f_5$: Request ($\mathcal{D}_5$) \\ 
    \midrule
    Unprepared for the attack, the Swedish attempted to \textbf{save$_{[1]}$} their ships by cutting their anchor ropes and to \textbf{flee$_{[2]}$}. & [1] Rescuing ($\mathcal{D}_2$) [2] Escaping ($\mathcal{D}_3$) & [1] \underline{$f_{2:4}$}: Rescuing; $f_5$: Other \hspace{15 mm}  [2] \underline{$f_{3:4}$}: Escaping; $f_5$: Other  \\
    \midrule
    After the uprising in Germany was \textbf{suppressed}, it flared briefly in several Swiss Cantons. & Control ($\mathcal{D}_3$) & $f_{3:5}$: Hindering ($\mathcal{D}_2$) \\
    \midrule
    Brazilians and Chinese living in the region have been \textbf{evacuated}. & Escaping ($\mathcal{D}_3$) & $f_{3:5}$: Removing ($\mathcal{D}_1$) \\
    \midrule
    A surveillance video of the incident was \textbf{released} by police four days after the shooting, on 26 November. & Releasing ($\mathcal{D}_3$) & $f_{3:4}$: Other; $f_{5}$: Publishing ($\mathcal{D}_5$)\\
    \midrule
    Giral \textbf{agreed} to arm the trade unionists in defence of the Republic, and had 60, 000 rifles delivered to the CNT and UGT headquarters, although only 5, 000 were in working order. & Agree\_or\_refuse\_to\_act ($\mathcal{D}_4$) & $f_{4}$: Other; \hspace{50 mm} \underline{$f_5$}: Agree\_or\_refuse\_to\_act \\
    \midrule
    Meanwhile, in the city, the Republican government had \textbf{reformed} under the leadership of socialist leader Francisco Largo Caballero. & Reforming\_a\_system ($\mathcal{D}_1$) & \underline{$f_{1,2,4,5}$}: Reforming\_a\_system; \hspace{13 mm} $f_{3}$: Change\_of\_leadership ($\mathcal{D}_3$) \\
    \bottomrule
    \end{tabular}
	} 
	\caption{Failure analysis of our EMP on the first permutation of MAVEN. The targeted triggers are highlighted in \textbf{bold}. $\mathcal{D}_i$ after the event types indicate the type is introduced at $i$-th task. $f_i$ indicates the model trained after $i$-th task. We highlight the models predicted the correct types with \underline{underline}. For example, "Coming\_to\_be ($\mathcal{D}_2$)" indicates the \emph{Coming\_to\_be} type is introduced at the 2nd task. "$f_{3:4}$: Other" indicates the models trained after the 3rd and 4th task both predict the \emph{Other} type. }
	\vspace{-1em}
	\label{tab:case_study}
\end{table*}

\end{document}